\documentclass[final]{l4dc2026}

\usepackage{mathtools}
\usepackage{enumitem}
\usepackage{algorithmic}
\usepackage{booktabs}
\usepackage{tablefootnote}
\usepackage{caption}

\title[Certified Robust Invariant Polytope Training]{Certified Robust Invariant Polytope Training in \\ Neural Controlled ODEs}
\usepackage{times}
\coltauthor{\Name{Akash Harapanahalli} \Email{aharapan@gatech.edu} \\ 
 \Name{Samuel Coogan} \Email{sam.coogan@gatech.edu}\\
 \addr School of Electrical and Computer Engineering, Georgia Institute of Technology}

\DeclareSymbolFont{bbold}{U}{bbold}{m}{n}
\DeclareSymbolFontAlphabet{\mathbbold}{bbold}

\newcommand{\smallconc}[2]{\begin{bsmallmatrix} #1 \\ #2 \end{bsmallmatrix}}

\newcommand{\geqse}{\geq_{\mathrm{SE}}}
\newcommand{\leqse}{\leq_{\mathrm{SE}}}

\renewcommand{\>}{\rangle}

\newcommand{\Tgeq}{\calT_{\geq0}}

\newcommand{\relu}{\operatorname{ReLU}}
\newcommand{\ox}{\mathring{x}}
\newcommand{\ou}{\mathring{u}}
\newcommand{\ow}{\mathring{w}}

\newcommand{\R}{\mathbb{R}}

\newcommand{\IR}{\mathbb{IR}}

\newcommand{\calH}{\mathcal{H}}
\newcommand{\calI}{\mathcal{I}}

\newcommand{\calL}{\mathcal{L}}

\newcommand{\calS}{\mathcal{S}}
\newcommand{\calT}{\mathcal{T}}

\newcommand{\calW}{\mathcal{W}}

\newcommand{\bff}{\mathbf{f}}

\newcommand{\bfu}{\mathbf{u}}

\newcommand{\bfw}{\mathbf{w}}
\newcommand{\bfx}{\mathbf{x}}

\newcommand{\bfI}{\mathbf{I}}

\newcommand{\bfW}{\mathbf{W}}

\newcommand{\sfE}{\mathsf{E}}
\newcommand{\sfF}{\mathsf{F}}
\newcommand{\sfG}{\mathsf{G}}

\newcommand{\sfM}{\mathsf{M}}

\newcommand{\ul}[1]{\underline{#1}}

\newcommand{\uld}{\ul{d}}

\newcommand{\ulu}{\ul{u}}

\newcommand{\ulw}{\ul{w}}
\newcommand{\ulx}{\ul{x}}
\newcommand{\uly}{\ul{y}}
\newcommand{\ulz}{\ul{z}}

\newcommand{\ulC}{\ul{C}}

\newcommand{\ulH}{\ul{H}}

\newcommand{\ol}[1]{\overline{#1}}

\newcommand{\old}{\ol{d}}

\newcommand{\olu}{\ol{u}}

\newcommand{\olw}{\ol{w}}
\newcommand{\olx}{\ol{x}}
\newcommand{\oly}{\ol{y}}
\newcommand{\olz}{\ol{z}}

\newcommand{\olC}{\ol{C}}

\newcommand{\olH}{\ol{H}}

\begin{document}

\maketitle

\begin{abstract}%
We propose a framework for training neural network controllers with certified robust forward invariant polytopes.
First, we parameterize a family of lifted control systems in a higher dimensional space, where the original neural controlled system evolves on an invariant subspace of each lifted system. 
We use interval analysis and neural network verifiers to further construct a family of lifted embedding systems, carefully capturing the knowledge of this invariant subspace.
If the vector field of any lifted embedding system satisfies a sign constraint at a single point, then a certain convex polytope of the original system is robustly forward invariant. 
Treating the neural network controller and the lifted system parameters as variables, we propose an algorithm to train controllers with certified forward invariant polytopes in the closed-loop control system.
Through two examples, we demonstrate how the simplicity of the sign constraint allows our approach to scale with system dimension to over $50$ states, and outperform state-of-the-art Lyapunov-based sampling approaches in runtime.
\end{abstract}


\section{Introduction} \label{sec:intro}

Learning-enabled components are increasingly used in closed-loop control systems due to their ease of computation and ability to outperform optimization-based feedback control approaches~\citep{SC-etal:18}. 
In safety-critical control systems, ensuring the reliability of these learning-enabled components is crucial for deployment.
Recent work has focused on verifying and training robust neural networks, as summarized in~\cite{VNN-COMP} and~\cite{meng2022adversarial}.
Neural networks in control loops introduce unique challenges such as the compounding of error via feedback, and verifying robustness in the closed-loop setting has also been the subject of recent work~\citep{ARCH24:ARCH_COMP24_Category_Report_Artificial}. 
However, there are few methods for training safe neural network feedback controllers, despite advancements in certified robust training of isolated neural networks.

Certifying a robust forward invariant set is a well-studied and natural technique for ensuring infinite-time safe behavior of systems. 
These safe sets can represent a variety of different physical specifications, including operating regions, goal regions, or the complement of unsafe regions (\emph{e.g.}, obstacles).
There are many classical techniques for computationally certifying invariance specifications, including Lyapunov-based analysis using sum-of-squares programming~\citep{AP-SP:02,UT-AP-PS:08}, barrier-based methods which introduce an online convex optimization problem to solve for each control input~\citep{AA-etal:2019}, and set-based approaches which require explicit characterizations of tangent cones~\citep{FB:99}. 
However, a direct application of these methods generally fails when facing high-dimensional and nonlinear neural network controllers in-the-loop. 

Robustness of neural networks in isolation is a well studied field in the machine learning community---for a recent survey, see~\cite{meng2022adversarial}.
A key feature of many of these approaches is their implementation supporting automatic differentiation, to help promote robustness during the training procedure.
As opposed to pure input-output robustness of neural networks, the challenge with closed-loop feedback is to capture the interactions of the network and the system. 
There is a growing community centered specifically around studying the safety of neural networks applied in feedback loops.
Approaches based on computing reachable sets of the neural controlled system include POLAR~\citep{CH-JF-XC-WL-QZ:22}, JuliaReach~\citep{CS-MF-SG:22}, NNV~\citep{HDT-etal:20}, CORA~\citep{NK-etal:23}, and ReachMM~\citep{SJ-AH-SC:24} for nonlinear systems, and %
ReachLP~\citep{ME-GH-CS-JPH:21} and %
Reach-SDP~\citep{HH-MF-MM-GJP:20} for linear systems. 
We refer to~\cite{ARCH24:ARCH_COMP24_Category_Report_Artificial} for a comprehensive list of benchmarks and tools the community has been studying.
However, to our knowledge, none of these approaches for control systems support autodifferentiation to help train neural network controllers with safety guarantees.

There are some papers that study forward invariance for neural networks in dynamical systems.
The paper \cite{AS-RGS:21} %
verifies invariant interval sets for control-affine systems with independent inputs for each state variable, \cite{LJ-AS-SO:23} finds invariant non-convex regions for linear systems with piecewise affine controllers, \cite{HY-PS-MA:22} finds an ellipsoidal inner-approximation of a region of attraction for the system using Integral Quadratic Constraints (IQCs). In~\cite{HD-BL-LY-MP-RT:21}, a Lyapunov-based approach is used to find robust invariant sets of control systems modeled by neural networks.  
For training robust neural ODEs, LyaNet~\citep{rodriguez2022lyanet} is a Lyapunov-based approach to improve stability and \cite{WX-etal:2023} uses control barrier functions to filter parameters online to ensure set invariance. 
For neural ODEs and neural network controlled systems,~\cite{huang2023fiode} sample along the boundary of a Lyapunov function to certify and train neural networks with robust invariance guarantees.

\paragraph*{Contributions}
We first propose a novel technique to address the problem of certifying robust forward invariant polytopes.
We introduce the lifted system, which is a lifting of the closed-loop system~\eqref{eq:clsys} into a $n$-dimensional subspace of $\R^m$ using a tall matrix $H$ and a parameterized left inverse $H^+$.
For any bounded and convex polytope, we construct a family of lifted embedding systems parameterized by $H^+$.
A component-wise positivity check on the vector field at a single point, for any instance of $H^+$, obtains a sufficient condition for robust forward invariance for the original neural network controlled system~\eqref{eq:clsys}.
Compared to previous approaches, our method does not require sampling along the entire boundary of a desired robust invariant set, improving certification runtime and scalability with respect to state dimension.
We then propose a novel method for training certified robust forward invariant polytopes by incorporating the positivity condition from the lifted embedding system into the optimization problem. 
A simple loss function induces the desired positivity in the lifted embedding space, which can subsequently be checked at each training iteration at little cost.
Next, we add an unconstrained decision variable $\eta$ by leveraging the parameterization of possible left inverses $H^+$, which reduces overconservatism by searching through the entire family of possible lifted dynamics.
We implement the proposed algorithm in JAX, which allows us to just-in-time compile and vectorize the embedding system evaluations onto a GPU.
The simplicity of our robust invariance condition allows our algorithm to demonstrate better training times and better scalability as compared to a previous sampling-based approach.

\paragraph*{Notation}
Define the partial ordering $\leq$ on $\R^n$ such that $\ulx\leq\olx\iff \ulx_i\leq\olx_i$ for every $i=1,\dots,n$. Let $[\ulx,\olx] := \{x\in\R^n : \ulx \leq x \leq \olx\}$ denote a closed and bounded interval in $\R^n$, and let $\IR^n$ denote the set of all such intervals. 
Define the upper triangle $\Tgeq^{2n} := \{\smallconc{\ulx}{\olx}\in\R^{2n} : \ulx \leq \olx\}$, and note that $\IR^n\simeq\Tgeq^{2n}$. We denote this equivalence with $\left[\smallconc{\ulx}{\olx}\right] := [\ulx,\olx]$.
The partial order $\leq$ on $\R^n$ induces the southeast partial order $\leq_{\text{SE}}$ on $\R^{2n}$, where $\smallconc{x}{\hat{x}}\leqse\smallconc{y}{\hat{y}}\iff x\leq y$ and $\hat{y}\leq\hat{x}$. 
For $x^1\in\R^{n_1}$, $x^2\in\R^{n_2}$, $\dots$, $x^m\in\R^{n_m}$, let $(x^1,x^2,\dots,x^m)\in\R^{n_1 + n_2 + \dots + n_m}$ denote their concatenation. 
For two vectors $x,y\in\R^n$ and $i\in\{1,\dots,n\}$, let $x_{i:y}
\in\R^n$ be the vector obtained by replacing the $i$th entry of $x$ with that of $y$, \emph{i.e.},  $(x_{i:y})_j=y_j$ if $i=j$ and otherwise $(x_{i:y})_j=x_j$.
Let $\bfI_n\in\R^{n\times n}$ denote the $n\times n$ identity matrix.
We represent a \emph{convex polytope} as a nonempty set $\<H,\uly,\oly\> := \{x\in\R^n : \uly \leq Hx \leq \oly\}$, for a full rank matrix $H\in\R^{m\times n}$.
\footnote{Under these assumptions, convex polytopes are bounded. We note that any compact H-rep convex polytope $\{x\in\R^n : Hx \leq y\}$ may be written as $\<H,\uly,y\>$ by taking, \emph{e.g.}, sufficiently small $\uly$.}

\paragraph*{Preliminaries}

Consider the following nonlinear control system with disturbance,
\begin{align} \label{eq:olsys}
    \dot{x} = f(x,u,w),
\end{align}
where $x\in\R^n$ is the state of the system, $u\in\R^p$ is the control input, $w\in\calW\subset\R^q$ is some disturbance in compact set $\calW$, and $f:\R^n\times\R^p\times\R^q\to\R^n$ is a locally Lipschitz vector field.
Let $\pi:\R^n\to\R^p$ denote a neural network controller for the system, and define the closed-loop system
\begin{align} \label{eq:clsys}
    \dot{x} = f^\pi (x, w) := f(x,\pi(x),w).
\end{align}
For the system~\eqref{eq:clsys}, let $[0,\infty)\ni t\mapsto\phi_{f^\pi}(t,x_0,\bfw)$ denote its unique trajectory from initial condition $x_0$ at time $0$ under piecewise continuous disturbance mapping $\bfw:[0,\infty)\to\calW$.
The set $\calS\subseteq \R^n$ is $\calW$-\emph{robustly forward invariant} if $x_0\in\calS$ implies that $\phi_{f^\pi}(t,x_0,\bfw)\in\calS$ for any $t\geq 0$ and any piecewise continuous $\bfw:[0,\infty)\to\calW$.
The goal of this work is to \emph{train} the neural network $\pi$ such that a given polytope $\<H,\uly,\oly\>$ is a $\calW$-robustly forward invariant set for the closed-loop system~\eqref{eq:clsys}.

\section{The Closed-Loop Embedding System}
One of the biggest challenges in verifying neural network controlled dynamical systems is bounding the nonlinear behavior of the neural network controller while capturing its stabilizing closed-loop effects, which are important for invariance analysis.
In this section, we recap the approach from~\cite{SJ-AH-SC:24}, which builds inclusion functions and embedding systems that capture the first-order interactions of the system with the neural network. 

\subsection{Closed-Loop Inclusion Function}

First, we discuss two computational tools we use to bound the closed-loop dynamics~\eqref{eq:clsys}.
The first tool is CROWN~\citep{HZ-etal:18}, which obtains a \emph{local affine bound}. Given an interval $[\ulx,\olx]$, CROWN propogates linear bounds to obtain a tuple $(\ulC,\olC,\uld,\old)$ satisfying
\begin{align*}
    \ulC x + \uld \leq \pi(x) \leq \olC x + \old,
\end{align*}
valid for every $x\in[\ulx,\olx]$.
The second tool is a \emph{mixed Jacobian-based inclusion} for the open-loop dynamics $f$ from~\eqref{eq:olsys}: given centering points and intervals $\ox\in[\ulx,\olx]$, $\ou\in[\ulu,\olu]$, $\ow\in[\ulw,\olw]$, there are interval mappings $\sfM_x,\sfM_u,\sfM_w$ each with argument $(\ulx,\olx,\ulu,\olu,\ulw,\olw)$ such that
\begin{align*}
    f(x,u,w) \in &\ [\sfM_x](x - \ox) +[\sfM_u](u - \ou) + [\sfM_w](w - \ow) 
    + f(\ox,\ou,\ow),
\end{align*}
for any $x\in[\ulx,\olx]$, $u\in[\ulu,\olu]$, $w\in[\ulw,\olw]$.
For instance, inclusion functions for the Jacobian matrices of the map with respect to $(x,u,w)$ imply the inclusion by the mean value theorem---however, we use the \emph {mixed Jacobian matrix} which is less conservative~\citep{harapanahalli2024linear}.
The toolbox \verb|immrax|~\citep{AH-SJ-SC:24} automatically constructs these bounds using automatic differentiation and interval analysis.
Given the open-loop system~\eqref{eq:olsys} and a neural network controller $\pi$, we next build a closed-loop mixed Jacobian-based \emph{inclusion function} as
\begin{gather} 
    \nonumber
\begin{aligned}
    \sfF^\pi(\ulx,\olx,\ulw,\olw) &= \bigg[\begin{smallmatrix}
        \ulH^+ - \ul{\sfM}_x & \ulH^- \\
        \olH^- - \ol{\sfM}_x & \olH^+ \\
    \end{smallmatrix}\bigg] \smallconc{\ulx}{\olx} +
    \begin{bsmallmatrix}
        -\ul{\sfM}_w^- & \ul{\sfM}_w^- \\
        -\ol{\sfM}_w^+ & \ol{\sfM}_w^+
    \end{bsmallmatrix} \smallconc{\ulw}{\olw} 
    + \smallconc{-\ul{\sfM}_u\ulu + \ul{\sfM}_u^+\uld_{[\ulx,\olx]} + \ul{\sfM}_u^-\old_{[\ulx,\olx]} + f(\ulx,\ulu,\ulw)}{-\ol{\sfM}_u\ulu + \ol{\sfM}_u^+\old_{[\ulx,\olx]} + \ol{\sfM}_u^-\uld_{[\ulx,\olx]} + f(\ulx,\ulu,\ulw)}, 
\end{aligned} \\
    \ulH = \ul{\sfM}_x + \ul{\sfM}_u^+ \ulC + \ul{\sfM}_u^- \olC, \quad
    \olH = \ol{\sfM}_x + \ol{\sfM}_u^+ \olC + \ol{\sfM}_u^- \ulC, 
    \label{eq:clinclfun}
\end{gather}
where $(A^+)_{ij} = \max\{A_{ij},0\}$ and $A^- = A - A^+$ for any matrix $A$.
As shown in~\cite[Thm. 3]{SJ-AH-SC:24}, this inclusion function satisfies the bound $f^\pi(x,w) \in \sfF^\pi(\ulx,\olx,\ulw,\olw)$ for every $x\in[\ulx,\olx]$ and $w\in[\ulw,\olw]$.
In general, this approach works for any choice of $(\ox,\ou,\ow)$ equal to a corner of the box $[\ulx,\olx]\times[\ulu,\olu]\times[\ulw,\olw]$, with slight modifications to the expression~\eqref{eq:clinclfun}.
As shown in~\cite{SJ-AH-SC:24}, the inclusion function~\eqref{eq:clinclfun} captures the first-order interactions between the dynamics and the neural network controller in the first term multiplying $\smallconc{\ulx}{\olx}$.

\subsection{Hyperrectangle Invariance Using the Embedding System}

The mixed Jacobian-based closed-loop inclusion function provides an efficient and scalable technique for bounding the output of the closed-loop vector field $f^\pi$ in~\eqref{eq:clsys}. The inclusion function~\eqref{eq:clinclfun} builds a \emph{closed-loop embedding system}, 
    \begin{align} \label{eq:embsys}
    \begin{aligned}
        \dot{\ulx}_i & = \big(\ul{\sfE}(\ulx,\olx,\ulw,\olw)\big)_i := \big(\ul{\sfF}^\pi(\ulx,\olx_{i:\ulx},\ulw,\olw)\big)_i, \\
        \dot{\olx}_i & = \big(\ol{\sfE}(\ulx,\olx,\ulw,\olw)\big)_i := \big(\ol{\sfF}^\pi(\ulx_{i:\olx},\olx,\ulw,\olw)\big)_i,
    \end{aligned}
    \end{align}
    where $\smallconc{\ulx}{\olx}\in\Tgeq^{2n}$, $\smallconc{\ulw}{\olw}\in\Tgeq^{2q}$, and $\sfE:\Tgeq^{2n}\times\Tgeq^{2q}\to\R^{2n}$.
The embedding system can be thought of as evolving each face of the hyperrectangle separately, in a manner that contains the behavior of the original dynamics.
The evaluation separately on each face ($[\ulx,\olx_{i:\ulx}]$ and $[\ulx_{i:\olx},\olx]$) of the hyperrectangle is a key feature of our approach.
In our Python implementation, we use JAX to vectorize these faces for efficient evaluation on a GPU.
The next Proposition is from~\cite[Prop. 1]{AH-SJ-SC:23c}, and describes how robust forward invariance is simplified to a single evaluation of the vector field~\eqref{eq:embsys}.

\begin{proposition}[Invariant hyperrectangles] \label{prop:invhyper}
    Consider the closed-loop system~\eqref{eq:clsys}, with the inclusion function $\sfF^\pi$ from~\eqref{eq:clinclfun}, and the induced embedding system $\sfE$ from~\eqref{eq:embsys}. If
    \[\sfE(\ulx_0,\olx_0,\ulw,\olw) \geqse 0,\]
    then the hyperrectangle $[\ulx_0,\olx_0]$ is $[\ulw,\olw]$-robustly forward invariant for the closed-loop system~\eqref{eq:clsys}. 
\end{proposition}

Recall that $\sfE(\ulx_0,\olx_0,\ulw,\olw) \geqse 0$ if and only if $\ul\sfE(\ulx_0,\olx_0,\ulw,\olw) \geq 0$ and $\ol\sfE(\ulx_0,\olx_0,\ulw,\olw) \leq 0$.
Proposition~\ref{prop:invhyper} characterizes a boundary condition on the hyperrectangle $[\ulx,\olx]$ based on the Nagumo theorem~\cite[Thm. 3.1]{FB:99}---namely, these conditions ensure that the vector field points inside of the box along the entire boundary $\partial[\ulx,\olx]$, thus, trajectories can never escape the box.

\section{Polytope Invariance Using the Lifted Embedding System}
The approach from the previous section is sufficient for forward invariant intervals of $\R^n$, \emph{i.e.}, axis-aligned hyperrectangles. 
However, there are many systems that do not admit interval invariant sets under any controller. 
For instance, the phenomenon illustrated in the following example is inherent to many second-order control systems, such as mechanical systems for which the control is applied as a force, which is integrated twice to give the position state.
In this section, we extend the theory to verify arbitrary compact polytopes.

\begin{example}[Mechanical systems] \label{ex:mechsys}
    Consider the following mechanical system
    \[\dot{x}_1 = x_2,\ \dot{x}_2 = u,\]
    where $x_1$ is the position, $x_2$ is the velocity, and $u$ is the applied force.
    In these coordinates, it is impossible to design a controller $u:=\pi(x)$ such that a hyperrectangle is forward invariant. To see this, consider the box $\calS_1 = [-1,1]\times[-1,1]$. The point $[1, 1]^T\in\calS_1$, and the vector field is $f([1, 1]^T) = [1, u]^T$. Regardless of the control, $\dot{x}_1>0$, and the system will leave $\calS_1$. A similar argument holds for any other nonempty interval around the origin.
    Next, consider $u:=\pi(x) = -2x_1 - 3x_2$, and the transformation $T=\left[\begin{smallmatrix}1 & 1 \\ -1 & -2\end{smallmatrix}\right]$. In transformed coordinates $y:=T^{-1}x$,
    \[\dot{y} = T^{-1}(A+BK)Ty = \left[\begin{smallmatrix} -1 & 0 \\ 0 & -2 \end{smallmatrix}\right]y.\]
    Since the system is diagonal with negative eigenvalues, the vector field always points towards the origin, thus any hyperrectangle containing the origin is forward invariant for this transformed system.
    For example, the box $[-\frac12,\frac12]\times[-\frac12,\frac12]$ is forward invariant for the transformed system, implying that
    $\calS_2 = \<T^{-1},\smallconc{-1/2}{-1/2},\smallconc{1/2}{1/2}\>$
    is forward invariant for the original system.
    The sets $\calS_1$ and $\calS_2$ are visualized in Figure~\ref{fig:example1} in blue and green respectively.
\end{example}

\subsection{Lifted System}
The \emph{lifted system} embeds the original system into a $n$-dimensional subspace of $\R^m$. 

\begin{definition}[Lifted system]
    Consider the closed-loop system~\eqref{eq:clsys}, and let $H\in\R^{m\times n}$ be a full rank matrix.
    Let $H^+\in\R^{n\times m}$ be any matrix satisfying $H^+ H = \bfI_n$.
    Then
    \begin{align} \label{eq:liftedsys}
        \dot{y} = g(y,w) := H f^\pi(H^+ y, w) = H f(H^+ y, \pi(H^+y), w),
    \end{align}
    with state $y:=Hx$,
    is the \emph{$(H,H^+)$-lifted system} of~\eqref{eq:clsys}.
\end{definition}

In the next Proposition, we parameterize the set of left inverses $H^+$, which will allow us to incorporate this matrix as an unconstrained decision variable in the training problem.
\begin{proposition}[Parameterization of left inverses]~\label{prop:parameterized-lifting}
    Let $H\in\R^{m\times n}$ be full rank. Let $N\in\R^{m\times(m-n)}$ be a basis spanning the left nullspace of $H$ and let $H^\dagger = (H^TH)^{-1}H^T$ be the Moore-Penrose Pseudoinverse of $H$. Then the following characterizes the set of matrices satisfying $H^+H = \bfI_n$:
    \begin{align*}
        \{H^+\in\R^{n\times m} : H^+ = H^\dagger + \eta N^T,\,\eta\in\R^{n\times(m-n)}\}.
    \end{align*}
\end{proposition}

A key property of the lifted system is that the original system lies on an invariant $n$-dimensional subspace of the lifted state space $\R^m$.
The invariance of $\calH$ will be incorporated as extra knowledge for building a good embedding system for the lifted system~\eqref{eq:liftedsys} in the next section.

\begin{proposition}[Invariant subspace] \label{prop:invsubspace}
    Consider the closed-loop system~\eqref{eq:clsys}, with the $(H,H^+)$-lifted system~\eqref{eq:liftedsys}. For any $x_0\in\R^n$ and piecewise continuous $\bfw:[0,\infty)\to\calW$, 
    \[ 
    H\phi_{f^\pi}(t,x_0,\bfw) = \phi_g(t,Hx_0,\bfw).
    \]
    Moreover, the linear subspace $\calH := \{Hx : x\in\R^n\}$ is forward invariant for the lifted system.
\end{proposition}

\subsection{Lifted Embedding System}

One approach is to simply embed the lifted system and obtain a valid embedding system for the lifted dynamics~\eqref{eq:liftedsys}. However, this would discard the \emph{a priori} knowledge that the original system lives on the invariant subspace from Proposition~\ref{prop:invsubspace}. 
A technique for incorporating this information was explored in~\cite{KS-JKS:17}, where a refinement operator was used in conjunction with model redundancies to improve interval reachable set estimates for dynamical systems.
The following Definition allows us to incorporate the knowledge that the original system lies on the subspace $\calH$, and continue with the efficient interval analysis framework previously developed. 
\begin{definition}[Interval refinement operator]\label{def:refinement}
    Let $\calH\subset\R^m$ be a subset. $\calI_\calH:\Tgeq^{2m}\to\Tgeq^{2m}$ is an \emph{interval refinement operator} on $\calH$ if for every $[\uly,\oly]\in\IR^m$,
    \[\calH \cap [\uly,\oly] \subseteq [\ul{\calI}_\calH(\uly,\oly),\ol{\calI}_\calH(\uly,\oly)] \subseteq [\uly,\oly].\]
\end{definition}
For the case where $\calH = \{Hx : x\in\R^n\}$ is a subspace, we can use the fact that the left null space of $H$ encodes $(m-n)$ constraints on $\R^m$ equivalently defining the subspace $\calH$.
Given a library of left null vectors $A\in\R^{N\times m}$, where $AH = 0$, the following defines a valid interval refinement operator,
\begin{gather} \label{eq:IH}
\left[\calI_\calH (\uly,\oly)\right]_j = [\uly_j,\oly_j] \bigcap_{A_{i,j\neq 0}} -\frac{1}{A_{i,j}}\sum_{k\neq i} A_{i,k} [\ulz_k,\olz_k] 
\end{gather}
This is essentially equivalent to $\calI_G$ from~\cite{KS-JKS:17} which defines an interval refinement operator with the explicit knowledge that $Mz = b$ for some matrices $M$ and $b$, and is further explored in~\cite{gould_automatic_2025}, where left null vectors are sampled in a structured manner to promote sparsity.
We provide a proof that~\eqref{eq:IH} provides a valid refinement operator on $\calH$ in the appendices.

\begin{figure}
    \centering
    \includegraphics[width=0.33\columnwidth]{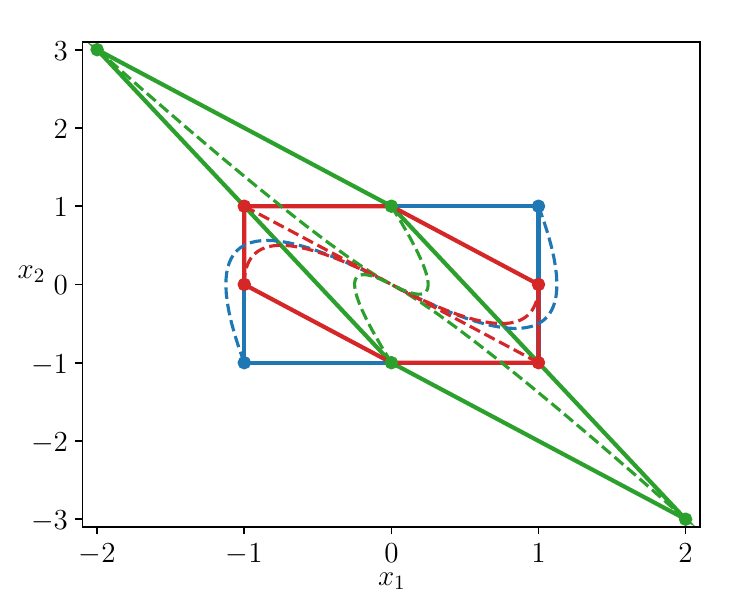}
    \includegraphics[width=0.66\columnwidth, clip]{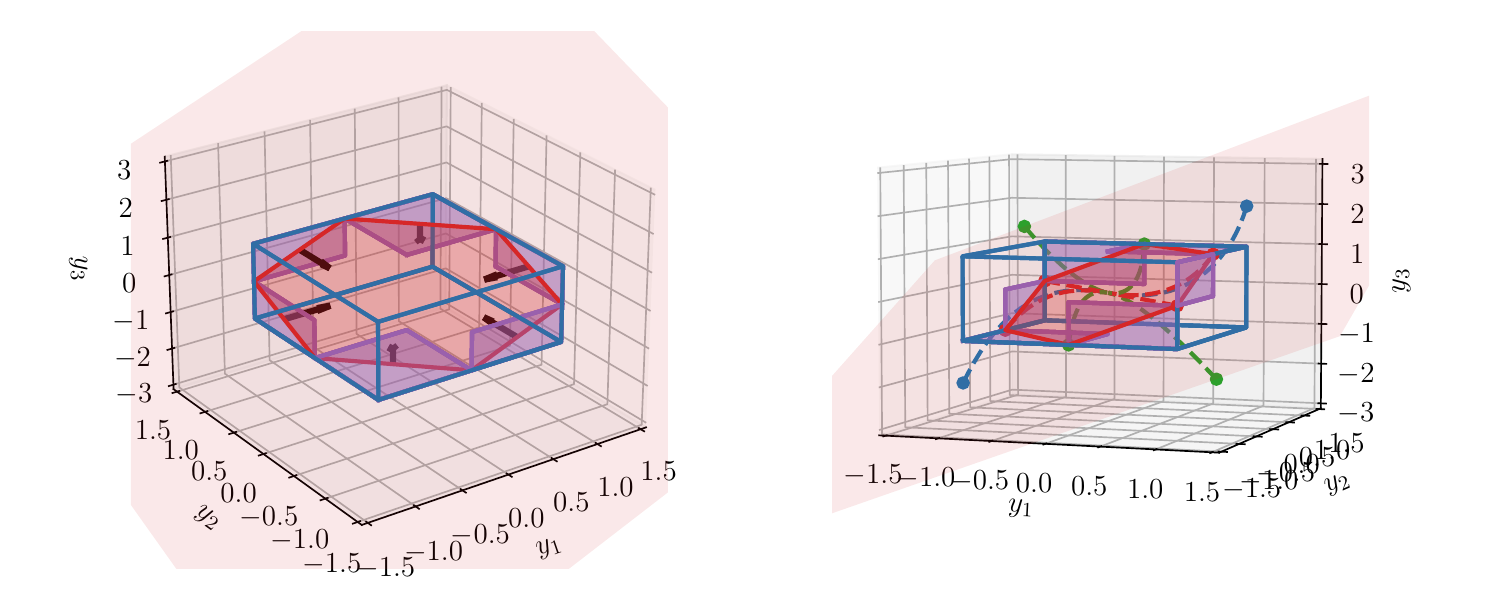}
    \caption{(\textbf{Left}) The mechanical system from Examples~\ref{ex:mechsys} and~\ref{ex:mechsyscont} is visualized with several polytopes in solid lines and solution trajectories in dotted lines. The blue set $\calS_1$ is a hyperrectangle that cannot be forward invariant, the green invariant set $\calS_2$ is obtained from the transformation associated with the eigenvalue decomposition, and the red invariant set $\calS_3$ is obtained by lifting the system into $3$-dimensions.
    (\textbf{Middle/Right}) $\calI_\calH$ is applied to every face of the box $[\uly,\oly] = [-1,1]^3$ (blue) for the subspace $\calH$ from Example~\ref{ex:mechsyscont} (red). The outputs (purple) are refined interval sets which still contain $\calH$. The red outlined set ($\calH\cap[\uly,\oly]$) corresponds to the polytope $\<H,\uly,\oly\>$ from Example~\ref{ex:mechsyscont} and the red polytope from the left figure.
    The original system evolves on the subspace $\calH$ by Proposition~\ref{prop:invsubspace}, and the positivity condition $\sfE_{H,H^+}(\uly,\oly,\ulw,\olw)\geqse0$ from Theorem~\ref{thm:polyreach} implies the vector component points in the direction indicated by the black arrows (\textbf{middle}) along each refined face. 
    The corresponding trajectories from the left figure are shown in dotted lines (\textbf{right}).
    }
    \label{fig:example1}
\end{figure}

\begin{definition}[Lifted embedding system]
Given a $\calH$-refinement operator $\calI_\calH$ and an inclusion function $\sfG$ for the lifted closed-loop dynamics~\eqref{eq:liftedsys}, define the $(H,H^+)$-lifted embedding system, 
\begin{align}\label{eq:liftedembsys}
    \begin{aligned} 
        \dot{\uly}_i &= \big(\ul{\sfG}(\ul{\calI}_\calH(\uly,\oly_{i:\uly}),\ol{\calI}_\calH(\uly,\oly_{i:\uly}),\ulw,\olw)\big)_i, \\
        \dot{\oly}_i &= \big(\ol{\sfG}(\ul{\calI}_\calH(\uly_{i:\oly},\oly),\ol{\calI}_\calH(\uly_{i:\oly},\oly),\ulw,\olw)\big)_i.
    \end{aligned}
\end{align}
\end{definition}
Since we know the original system lives on $\calH$ by Proposition~\ref{prop:invsubspace}, each face of the $[\uly,\oly]$ hyperrectangle can be refined to a smaller interval containing the intersection of $\calH$ and the face.
Figure~\ref{fig:example1} demonstrates this procedure for the $6$ faces of a $3$ dimensional hyperrectangle intersecting a $2$ dimensional subspace from Example~\ref{ex:mechsyscont} below.

The following Theorem describes how, once again, robust forward polytope forward invariance is simplified to a single evaluation of the vector field~\eqref{eq:liftedembsys}. The full proof is in the appendices.

\begin{theorem}[Polytope invariant sets] \label{thm:polyreach}
    Consider the closed-loop system~\eqref{eq:clsys}.
    Let $H\in\R^{m\times n}$, and $H^+\in\R^{n\times m}$ satisfy $H^+H = \bfI_n$. Let $\sfE_{H,H^+}$ denote the $(H,H^+)$-lifted embedding system~\eqref{eq:liftedembsys}.
    If
    \[\sfE_{H,H^+}(\uly,\oly,\ulw,\olw) \geqse 0,\]
    then the polytope $\<H,\uly,\oly\>$ is $[\ulw,\olw]$-robustly forward invariant for the original system~\eqref{eq:clsys}.
\end{theorem}

\begin{remark}[Comparison to the literature]
Theorem~\ref{thm:polyreach} generalizes \cite[Thm. 2]{AH-SJ-SC:23c}, which verified invariant polytopes when $H$ is square. (taking $H^+ = H^{-1}$ recovers the result).
\end{remark}
Similar to the original embedding system, the lifted embedding system provides a scalable and trainable condition for checking the forward invariance of a polytope.
In the next Example, we return to the mechanical system to demonstrate how this condition can be used to certify invariance.

\begin{example}[Mechanical system, cont.] \label{ex:mechsyscont}
    Consider the mechanical control system from Example~\ref{ex:mechsys}, with the same feedback controller. With the definitions 
    $H = \begin{bsmallmatrix}
        1 & 0 \\ 0 & 1 \\ 1 & 1
    \end{bsmallmatrix}$, $\uly = \begin{bsmallmatrix} -1 \\ -1 \\ -1 \end{bsmallmatrix}$, $\oly = \begin{bsmallmatrix} 1 \\ 1 \\ 1 \end{bsmallmatrix}$, $\calI_\calH$ from~\eqref{eq:IH}, and $\sfE_{H,H^\dagger}$ as the $(H,H^\dagger)$-lifted embedding system~\eqref{eq:liftedembsys}, 
    \[\sfE_{H,H^\dagger}(\uly,\oly) = [0,1,\tfrac43,0,-1,-\tfrac43 ]^T \geqse 0,\]
    thus the polytope $\calS_3 = \<H,\uly,\oly\>$ is a forward invariant set for the original system. 
    The polytope $\calS_3$ is visualized in Figure~\ref{fig:example1} in green.
    Additionally, the box $[\uly,\oly]$, subspace $\calH = \{Hx : x\in\R^2\}$, intersection $\calH\cap[\uly,\oly]$, and outputs of $\calI_\calH(\uly,\oly_{i:\uly})$ and $\calI_\calH(\uly_{i:\oly},\oly)$ are all visualized in Figure~\ref{fig:example1}.
    The black arrows show the direction ($+/-$) of the embedding vector field along those components, showing how the polytope is certified to be forward invariant using Theorem~\ref{thm:polyreach}.
\end{example}

\section{Certified Polytope Invariance Training}

Using the lifted embedding system, in this section we construct a loss for training controllers with certified forward invariant polytopes.
First, we assume there already exists an original loss $\calL^\text{data}$.
Ideally, we would add the positivity condition from Theorem~\ref{thm:polyreach} as a hard constraint, however, the complexity of neural networks often necessitates the use of unconstrained optimization algorithms. 
Instead, we use regularizing loss term, with the hope that the algorithm will eventually tend towards network parameters that evaluate as $\geqse 0$.
For a desired polytope $\calS=\<H,\uly,\oly\>$,
we use the loss
\begin{align}\label{eq:safeloss}
\begin{aligned}
    \calL(\pi,\eta) &= \calL^{\text{data}}(\pi) + \lambda\calL^\calS (\pi,\eta), \\
    \calL^{\calS}(\pi,\eta) &= \sum_{i=1}^m \relu(\ol{\sfE}_{H,H^+_\eta}(\uly,\oly,\ulw,\olw)_i + \varepsilon) 
    + \sum_{i=1}^m \relu(-\ul{\sfE}_{H,H^+_\eta}(\uly,\oly,\ulw,\olw)_i + \varepsilon)
\end{aligned}
\end{align}
to induce $[\ulw,\olw]$-robust forward invariance,
where $\sfE_{H,H^+_\eta}$ is the $(H,H^+_\eta)$-lifted embedding system of $f^\pi$, $\eta\in\R^{n\times(m-n)}$ is a decision variable choosing $H^+$ according to Proposition~\ref{prop:parameterized-lifting}, and $\varepsilon>0$ is a small numerical constant.
The $\relu$ avoids incurring negative loss when the constraint is satisfied, instead switching to purely minimizing the error to the data, allowing the model to improve its efficacy. 
Then, if the descent brings the optimization to a point where $\sfE_{H,H^+_\eta}\not\geqse 0$, the loss~\eqref{eq:safeloss} appears again. 
As a result, large values of $\lambda$ work well in practice.

\begin{remark}[Choice of $\eta$]
    While the choice of $\eta$ and $H^+_\eta$ may seem inconsequential, its inclusion as a parameter is empirically crucial in choosing a lifted system for invariance analysis.
    Analyzing~\eqref{eq:liftedsys}, the choice of $H^+$ changes the dynamics of the lifted system off of the invariant subspace $\calH$, which can drastically reduce the overconservatism of the lifted embedding system in practice.
\end{remark}

\section{Experiments}
\footnote{Experiments were performed on a computer running Kubuntu 22.04 with Intel Xeon Gold 6230, NVIDIA Quadro RTX 8000, and 64 GB of RAM. All code for the experiments is available at \url{https://github.com/gtfactslab/Polytope-Training}. For all of the experiment details, please see the appendices.}
We use~\verb|jax_verify| and \verb|immrax|~\citep{AH-SJ-SC:24} to compute the embedding system~\eqref{eq:embsys}.
JAX~\citep{jax2018github} vectorizes the evaluations on each face of the lifted hyperrectangle from Theorem~\ref{thm:polyreach} onto the GPU, Equinox~\citep{equinox_kidger} helps autodifferentiate~\eqref{eq:safeloss} for gradient evaluations, and Optax~\citep{deepmind2020jax} provides the gradient-based optimizer.

\paragraph*{Segway Model}

Consider the nonlinear dynamics of a segway from~\cite{segway},
\begin{align} \label{eq:segway}
    \begin{aligned}
        \begin{bmatrix}
            \dot{\phi} \\ \dot{v} \\ \ddot{\phi}
        \end{bmatrix} = \begin{bmatrix}
            \dot{\phi} \\
            \frac{\cos\phi(-1.8u + 11.5v+9.8\sin\phi)-10.9u+68.4v-1.2\dot{\phi}^2\sin\phi}{\cos\phi-24.7} \\
            \frac{(9.3u-58.8v)\cos\phi + 38.6u - 234.5v - \sin\phi(208.3+\dot{\phi}^2\cos\phi)}{\cos^2\phi - 24.7}
        \end{bmatrix}
    \end{aligned}
\end{align}
with state $x=[\phi\ v\ \dot{\phi}]^T\in\R^3$. To compare with the literature, we mimic the training procedure from~\cite{huang2023fiode}, where the network is trained to imitate the LQR gains from the linearization of the system around the origin, while certifying a robust forward invariant region around the equilibrium. 
The robustness is with respect to a $\pm2\%$ uncertainty in each of the system parameters, which we represent as a bounded multiplicative disturbance $(1+w_k)$ for $w\in[-0.02,0.02]^{11}$ applied independently to each system parameter from~\eqref{eq:segway}.

One approach to define a suitable polytope is to attempt to diagonalize the system, as demonstrated in Example~\ref{ex:mechsys} where a diagonalizing transformation yielded forward invariance since all eigenvalues were negative. This intuition extends to the nonlinear neural network controlled system: (i) let $A_\text{cl}$ be the Jacobian matrix of the linearized system in closed-loop with the LQR gains, \emph{i.e.}, $A_\text{cl} = \frac{\partial f}{\partial x}(0,0) + \frac{\partial f}{\partial u} (0,0) K$; 
(ii) let the Jordan decomposition of this matrix be $T^{-1}A_{cl}T = \Lambda$, where $\Lambda$ is in Jordan form;
(iii) choose the matrix $H = T^{-1}$, and fix offsets, \emph{e.g.}, $[\uly,\oly] = [-2,2]^3$.

\begin{figure}
    \centering
    \begin{minipage}[b]{0.40\columnwidth}
        \centering 
        \vspace{-1em}
        \includegraphics[width=0.8\columnwidth, trim={0 0 0 1cm}, clip]{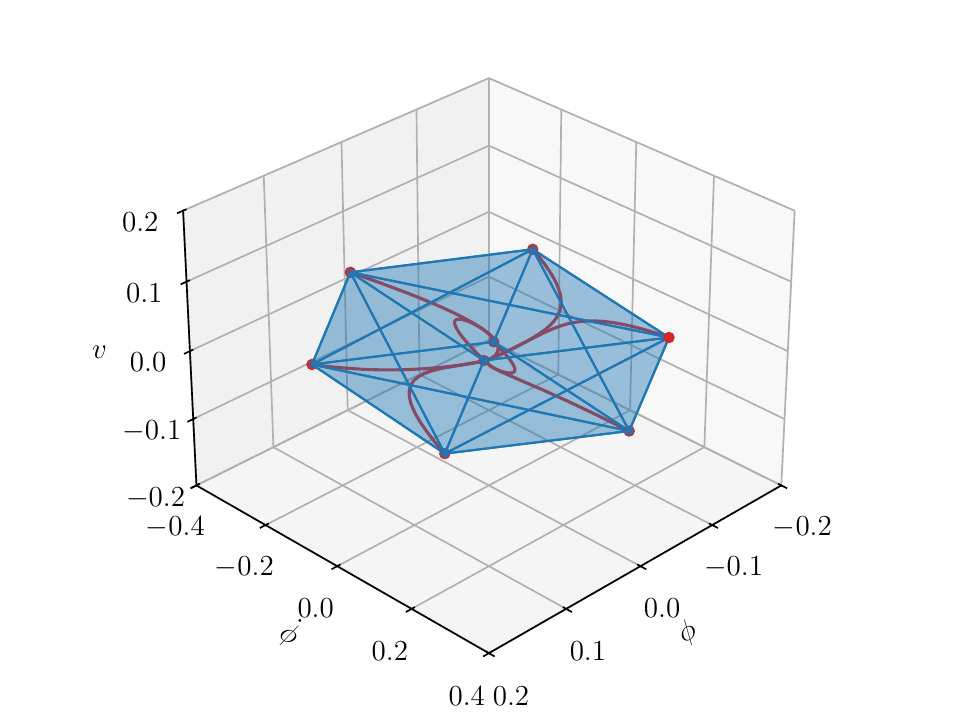}
        \captionof{figure}{The certified robust invariant polytope in $\R^3$ for the segway is visualized in blue. Simulations of trajectories starting from its vertices are in red.}
        \label{fig:segway}
    \end{minipage}
    \hfill
    \begin{minipage}[b]{0.58\columnwidth}
        \centering
        \vspace{-1em}
        \begin{tabular}{c|c|ccc}
            \toprule
              &  & \multicolumn{3}{c}{Runtime (s)} \\
            Method & Volume & Setup (JIT) & Training & Total \\
            \midrule
            \textbf{Ours} & 0.00152 & 139. & \textbf{11.4} & \textbf{150.4} \\
            FI-ODE        & \textbf{0.158} & \textbf{--} & 2758\tablefootnote{The time reported here includes both the training step (139 s) and the verification step (2619 s).} & 2758 \\
            \bottomrule
        \end{tabular}
        \captionof{table}{Comparison to robust invariance training literature}
        \label{tab:polytopecompare}
    \end{minipage}
\end{figure}

We compare to FI-ODE~\citep{huang2023fiode}, which to our knowledge, is the only other work that trains and certifies robust forward invariant sets in neural network controlled systems.
Our method requires an initial setup time to just-in-time (JIT) compile the optimizer step.
The positivity check in Theorem~\ref{thm:polyreach} verifies forward invariance in a fraction of the time compared to the sampling-based approaches by vectorizing over the faces of the hypercube in the lifted embedding space---allowing us to incorporate it directly into the objective. 
After compilation, it takes $11.4$ seconds to train a robust neural network controller, using ADAM~\citep{kingma2014adam} with a step size of $0.001$. 
The robust forward invariant polytope and sample system trajectories are visualized in Figure~\ref{fig:segway}.

The cost of our simple condition is possible overconservatism, demonstrated in our smaller volume compared with~\cite{huang2023fiode} in Table~\ref{tab:polytopecompare}.
We suspect this is because FI-ODE incorporates the $P$ matrix from their Lyapunov function into the initial training procedure, allowing them to shape the invariant set during training, while in this work we do not allow any shaping of the polytopes.
However, FI-ODE trains the neural network first without any guarantees, then a post-training sampling-based robust verification step verifies that a sublevel set of the Lyapunov function is robustly forward invariant, which suffers when scaling to higher dimensions.

\paragraph*{Platoon of Vehicles with Nonlinearities and Disturbances}

\begin{figure}
    \centering
    \begin{minipage}[b]{0.4\columnwidth}
        \centering 
        \vspace{-1em}
        \includegraphics[width=0.8\columnwidth]{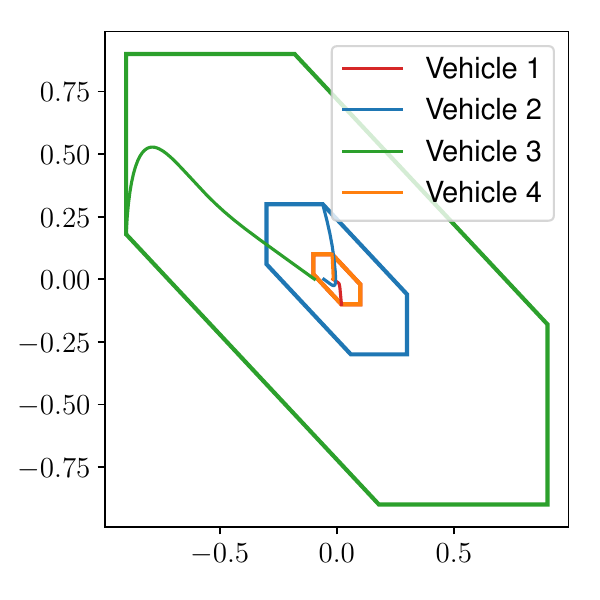}
        \captionof{figure}{The invariant polytope and a sample system trajectory for the platoon with $4$ vehicles is pictured. Vehicles $1$ and $4$ share the same invariant set in orange.}
    \end{minipage}
    \hfill
    \begin{minipage}[b]{0.58\columnwidth}
        \centering
        \vspace{-1em}
        \includegraphics[width=0.6\linewidth]{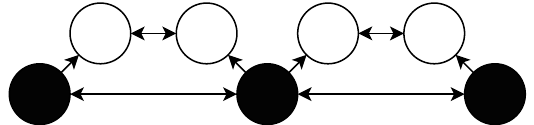}
        \captionof{figure}{The connection topology of the platoon. Leaders are filled and arrows represent relative measurements.}
        \label{fig:platoon_com}
        \begin{tabular}{c|cc|cc}
            \toprule
            & \multicolumn{2}{c|}{\# States} & \multicolumn{2}{c}{Runtime (s)} \\
            $N$ & Original & Lifted & Setup & Training (\#iter) \\
            \midrule
             4 &  8 & 12 & 35.8 & 6.90 (724) \\
            10 & 20 & 30 & 64.4 & 57.7 (725) \\
            16 & 32 & 48 & 128. & 170. (807) \\
            22 & 44 & 66 & 264. & 408. (890) \\
            28 & 56 & 84 & 478. & 1040 (1267) \\
            \bottomrule
        \end{tabular}
        \captionof{table}{Scalability with respect to number of vehicles}
        \label{tab:platoon}
    \end{minipage}
\end{figure}

In this example, we investigate how our proposed approach scales with state dimension $n$.
We consider a platoon of $N$ vehicles, each with the following dynamics,
\begin{gather} \label{eq:platoondyn}
    \dot{p}_j = v_j, \quad \dot{v}_j = \sigma(u_j)(1 + w_j),
\end{gather}
where for each vehicle $j=1,\dots,N$, $p_j\in\R$ is its position, $v_j$ is its velocity, $u_j\in\R$ is its control input, 
$w_j\in[-0.1,0.1]$ is a bounded disturbance input, and $\sigma(u) = u_\mathrm{lim}\tanh(u/u_\mathrm{lim})$ is a softmax nonlinearity, $u_\mathrm{lim} = 10$. 
Let $x_j = [p_j, v_j]^T\in\R^2$ for each $j$.
Each vehicle is controlled by a shared neural network control policy $\pi:\R^2\times\R^2\times\R^2\to\R$, 
\begin{align} \label{eq:platoonpolicy}
u_j = 
\begin{cases}
    \pi((x_j,\ x_{j-3} - x_{j},\ x_{j} - x_{j+3})), & j=3k, \\
    \pi((0,\ x_{j-1} - x_j,\ x_j - x_{j+1})), & \text{otherwise},
\end{cases}
\end{align}
with $x_0 := 0$, $x_{N+1} := 0$.
Every $3$rd vehicle (leader) measures its true state, and the relative difference between the next two leaders. The rest of the platoon (followers) measures relative states between their nearest two neighbors.
This communication scheme is pictured in Figure~\ref{fig:platoon_com}.
The closed-loop system can be rewritten as
$\dot{\bfx} = \bff(\bfx,\Pi(\bfu),\bfw) = \bff^{\Pi} (\bfx,\bfw)$,
where $\bfx = (x_1,\dots,x_N)\in\R^{2N}$, $\bfu = (u_1,\dots,u_N)\in\R^N$, $\bfw = (w_1,\dots,w_N)\in\R^N$, $\bff:\R^{2N}\times\R^N\times\R^N\to\R^{2N}$ is the dynamics~\eqref{eq:platoondyn}, $\Pi:\R^N\to\R^N$ represents the policy~\eqref{eq:platoonpolicy}. 

We would like to train the shared feedback policy $\pi$ such that the closed-loop platoon renders the polytope $\<H,\uly,\oly\>$, for $H = \bfI_N \otimes \begin{bsmallmatrix} 1 & 0 \\ 0 & 1 \\ 1 & 1 \end{bsmallmatrix}$, $\oly = \underbrace{[1,3,9,1,3,9,\dots,1,3,9,1]^T}_{\in\R^N} \otimes [0.1,0.1,0.08]^T$, $\uly = -\oly$, robustly forward invariant.
We use the loss~\eqref{eq:safeloss} with no data loss ($\calL^\mathrm{data} = 0$), $\lambda = 1$, a $6\times32\times32\times32\times1$ fully connected ReLU network for $\pi$, and a range of different numbers of vehicles $N$.
The set $\<H,\uly,\oly\>$ is illustrated in Figure~\ref{fig:segway}.
In Table~\ref{tab:platoon}, we outline how the training time scales with the number of state dimensions of the system and the lifted system.
Compared to sampling based approaches which suffer from the curse of dimensionality, our approach scales reasonably in both setup and training time.

\section{Conclusions}
In this paper, we proposed a framework for training certified robust forward invariant polytopes in neural network controlled dynamical systems, using a novel lifted embedding system where a single evaluation certifies forward invariance.
Through two experiments, we demonstrated how our approach both improves on existing sampling-based approaches in runtime, and scales well with state dimension.
In future work, we plan to address the overconservatism of our approach by incorporating the polytope itself into the optimization.

\section*{ACKNOWLEDGMENT}

This work was supported in part by the National Science Foundation under awards \#2219755, \#2333488, and \#2440387, and by the Air Force Office of Scientific Research under Grant FA9550-23-1-0303.

\bibliography{citations}
\newpage
\appendix
\onecolumn

\section{Algorithm for Certified Robust Forward Invariant Neural Network Training}

The following Algorithm describes how to use the loss function~\eqref{eq:safeloss} to promote forward invariance, and how to use the condition from Theorem~\ref{thm:polyreach} to easily check for certified forward invariance.

\begin{algorithm}[h]
   \caption{Certified polytope training using the lifted embedding system}
   \label{alg:training}
\begin{algorithmic}
    \STATE \textbf{Input:} Desired polytope $\<H,\uly,\oly\> = \{x\in\R^n : \uly \leq Hx \leq \oly\}$, $H\in\R^{m\times n}$ full rank, $\uly,\oly\in\R^m$, regularization constant $\lambda \geq 0$, disturbance set $[\ulw,\olw]$, data loss $\calL^\mathrm{data}$
    \STATE $N \gets \operatorname{null}(H^T)$
    \STATE $\eta \gets \mathbf{0}^{(m-n)\times m}$
    \REPEAT
    \STATE $H^+_\eta\gets H^\dagger + \eta N^T$
    \STATE $\sfE \gets (H,H^+_\eta)\text{-Lifted Embedding System~\eqref{eq:liftedembsys} for~\eqref{eq:clsys}}$
    \STATE $\calL \gets \calL^\mathrm{data}(\pi) + \lambda \calL^\calS(\pi,\eta) $
    \STATE $(\pi,\eta)\gets\operatorname{step\_optimizer}(\nabla_{(\pi,\eta)}\calL)$
    \UNTIL{$\sfE(\uly,\oly,\ulw,\olw) \geqse 0$}
    \STATE \textbf{Return:} Neural network controller $\pi$ with certified forward invariant polytope $\<H,\uly,\oly\>$ for~\eqref{eq:clsys}.
\end{algorithmic}
\end{algorithm}

\section{Proofs of Main Results}

\subsection{Proof of Proposition~\ref{prop:parameterized-lifting} - Parameterization of left inverses}
\textbf{Statement:}  Let $H\in\R^{m\times n}$ be full rank. Let $N\in\R^{m\times(m-n)}$ be a basis spanning the left nullspace of $H$ and let $H^\dagger = (H^TH)^{-1}H^T$ be the Moore-Penrose Pseudoinverse of $H$. Then the set
\begin{align}
    \{H^+\in\R^{n\times m} : H^+ = H^\dagger + \eta N^T,\,\eta\in\R^{n\times(m-n)}\}
\end{align}
characterizes the set of matrices satisfying $H^+H = \bfI_n$.

\begin{proof}
Let $A =\{H^+\in\R^{n\times m} : H^+ = H^\dagger + \eta N^T,\,\eta\in\R^{n\times(m-n)}\}$, and let $B = \{H^+\in\R^{n\times m} : H^+H = \bfI_n\}$.

\noindent ($\subseteq$) Consider any $\eta\in\R^{n\times(n-m)}$, and let $H^+_\eta = H^\dagger + \eta N^T$. For any $x\in\R^n$,
\begin{align}
    H^+_\eta Hx = ((H^TH)^{-1}H^T + \eta N^T) Hx = (H^TH)^{-1}H^THx + \eta N^THx = x + \eta \mathbf{0}x = x,
\end{align}
since $(H^TH)^{-1}H^TH = \bfI_n$, and since $N$ is a basis for the left null-space of $H$. Thus, $A \subseteq B$.
\newline

\noindent ($\supseteq$) Let $H^+$ be any matrix satisfying $H^+H = \bfI_n$. Therefore, for any $x\in\R^n$,
\begin{align}
    H^+Hx = x \iff H^\dagger Hx + (H^+ - H^\dagger)Hx = x \iff (H^+ - H^\dagger)Hx = 0,
\end{align}
which implies that each row $h_j\in\R^m$, $h_j := (H^+ - H^\dagger)_j$ is in the left nullspace of $H$. Since $N$ is a basis for the left nullspace, for every $j=1,\dots,n$, there exists $\eta_j\in\R^{(m-n)}$ such that $N\eta_j = h_j$. Finally, with $\eta = [\eta_1\cdots\eta_j\cdots\eta_n]^T$,
\[(H^+ - H^\dagger) = \eta N^T,\]
which implies that $H^+ = H^\dagger + \eta N^T$. Thus, $B\subseteq A$.
\end{proof}

\subsection{Proof of Proposition~\ref{prop:invsubspace} - Invariant subspace}
\textbf{Statement:} 
    Consider the closed-loop system~\eqref{eq:clsys}, with the $(H,H^+)$-lifted system~\eqref{eq:liftedsys}. For any $x_0\in\R^n$ and piecewise continuous $\bfw:[0,\infty)\to\calW$, 
    \[ H\phi_{f^\pi}(t,x_0,\bfw) = \phi_g(t,Hx_0,\bfw) \quad \text{ and } \quad \phi_{f^\pi}(t,x_0,\bfw) = H^+\phi_g(t,Hx_0,\bfw). \]
    Moreover, the linear subspace $\calH := \{Hx : x\in\R^n\}$ is forward invariant for the lifted system.

\begin{proof}
Let $t\mapsto x(t)$ be the trajectory of the closed-loop system~\eqref{eq:clsys} under piecewise continuous $t\mapsto \bfw(t)$, \emph{i.e.}, $\phi_{f^\pi}(t,x_0,\bfw)$. Let $t\mapsto y(t)$ be the trajectory of the $(H,H^+)$-lifted system~\eqref{eq:liftedsys} from initial condition $Hx_0$ under $\bfw$, \emph{i.e.}, $\phi_g(t,Hx_0,\bfw)$.

Let $\hat{y}(t) := Hx(t)$ for every $t\geq 0$, which in turn implies that $H^+\hat{y}(t) = x(t)$. Note that
\begin{align}
    \dot{\hat{y}}(t) &= H\dot{x}(t) = Hf(x(t), w(t)) = Hf(H^+\hat{y}(t), w(t)),
\end{align}
which is the same dynamics as the $(H,H^+)$-lifted system~\eqref{eq:liftedsys}. Additionally, note that $\hat{y}(0) = Hx(0) = Hx_0$. Thus, since $y(t)$ and $\hat{y}(t)$ have the same dynamics and the same initial condition, and $f$ was assumed to be locally Lipschitz, the uniqueness of solutions to ODEs implies that $y(t) = \hat{y}(t)$. Thus, for every $t\geq 0$,
\begin{align*}
    y(t) = Hx(t), \text{ which also implies that }\ H^+ y(t) = x(t).
\end{align*}
Moreover, we have shown that for any initial condition $x_0\in\R^n$, $y_0 = Hx_0\implies y(t) = Hx(t)$ for every $t\geq 0$, which implies that the linear subspace $\calH=\{Hx:x\in\R^n\}$ is forward invariant for the lifted system~\eqref{eq:liftedsys}.
\end{proof}

\subsection{Proof of Interval Refinement Operator - $\calI_\calH$ implementation} \label{apx:IH}
\textbf{Need to show:} Given $AH = 0$, $\calI_\calH$ from~\eqref{eq:IH}
\begin{gather*} 
\left[\calI_\calH (\uly,\oly)\right]_j = [\uly_j,\oly_j] \bigcap_{A_{i,j\neq 0}} -\frac{1}{A_{i,j}}\sum_{k\neq i} A_{i,k} [\ulz_k,\olz_k] 
\end{gather*}
satisfies
\[\calH \cap [\uly,\oly] \subseteq [\ul{\calI}_\calH(\uly,\oly),\ol{\calI}_\calH(\uly,\oly)] \subseteq [\uly,\oly].\]

\begin{proof}
Let $H\in\R^{m\times n}$ be a full rank matrix, and let $A\in\R^{N\times m}$ satisfy $AH = 0$. Thus, $AHx = 0$ for any $x\in\R^n$, and for any $y\in\calH=\{Hx : x\in\R^n\}$,
\[Ay = 0 \implies \sum_{k=1}^{m} A_{i,k} y_k = 0 \text{ for every } i=1,\dots,m-n,\]
as the equation $Ay=0$ is the same as the system of equalities. Thus, for every $i=1,\dots,N$, and every $j=1,\dots,m$,
\[y_j = -\frac{1}{A_{i,k}}\sum_{k\neq j}A_{i,k}y_k.\]
With the additional information that $y\in[\uly,\oly]$, interval analysis on the RHS is still an overapproximation of $[\uly_j,\oly_j]$, but may provide a better overestimate. If not, the intersection with the original $[\uly_j,\oly_j]$ ensures the right inclusion.
Thus, $\calH\cap[\uly,\oly]\subseteq[\ul{\calI}_\calH(\uly,\oly),\ol{\calI}_\calH(\uly,\oly)] \subseteq [\uly,\oly]$.
\end{proof}

\subsection{Proof of Theorem~\ref{thm:polyreach} - Polytope invariant sets}
\textbf{Statement:}
    Consider the closed-loop system~\eqref{eq:clsys}.
    Let $H\in\R^{n\times m}$, and $H^+\in\R^{m\times n}$ satisfy $H^+H = \bfI_n$. Let $\sfE_{H,H^+}$ denote the $H,H^+$-lifted embedding system~\eqref{eq:liftedembsys}.
    If 
    \[\sfE_{H,H^+}(\uly,\oly,\ulw,\olw) \geqse 0,\]
    then the polytope $\<H,\uly,\oly\>$ is $[\ulw,\olw]$-robustly forward invariant for the original system.
\begin{proof}
By the equivalence from Proposition~\ref{prop:invsubspace}, it suffices to show that $\calH\cap[\uly,\oly]$ is $[\ulw,\olw]$-robustly forward invariant for the lifted system. 
Indeed, assume that $\calH\cap[\uly,\oly]$ is a $\calW$-robustly forward invariant set for the lifted system~\eqref{eq:liftedsys}. Let $t\mapsto \bfw(t)\in\calW$ be any piecewise continuous disturbance curve, and let $y_0\in\calH\cap[\uly,\oly]$. Since $y_0\in\calH$, $y_0 = Hx_0$ for some $x_0$. Since $\calH\cap[\uly,\oly]$ is $\calW$-robustly forward invariant, $\phi_g(t,Hx_0,\bfw)\in\calH\cap[\uly,\oly]$ for every $t\geq 0$. By Proposition~\ref{prop:invsubspace}, this implies that $H\phi_f(t,x_0,\bfw)\in\calH\cap[\uly,\oly]$ for every $t\geq 0$. But, since $Hx\in\calH$ for any $x\in\R^n$, this is the same as
\[\uly\leq H\phi_f(t,x_0,\bfw)\leq \oly \iff \phi_f(t,x_0,\bfw)\in \<H,\uly,\oly\>,\]
for every $t\geq 0$.

It remains to show that $\calH\cap[\uly,\oly]$ is indeed $[\ulw,\olw]$-robustly forward invariant.
Let $[\uly,\oly]$ satisfy that $\sfE_{H,H^+}(\uly,\oly,\ulw,\olw)\geqse 0$. Therefore, for every $i=1,\dots, m$,
\[ 0 \leq \ul{\sfG}_i(\ul{\calI}_\calH(\uly,\oly_{i:\uly}),\ol{\calI}_\calH(\uly,\oly_{i:\uly}),\ulw,\olw).\]
Since $\calI_\calH$ is a refinement operator,
\[\calH\cap[\uly,\oly_{i:\uly}] \subseteq [\ul{\calI}_\calH(\uly,\oly_{i:\uly}), \ol{\calI}_\calH(\uly,\oly_{i:\uly})],\]
therefore, it follows that
\[ 0 \leq \ul{\sfG}_i(\ul{\calI}_\calH(\uly,\oly_{i:\uly}),\ol{\calI}_\calH(\uly,\oly_{i:\uly}),\ulw,\olw) \leq \inf_{y\in\calH\cap[\uly,\oly_{i:\uly}],w\in[\ulw,\olw]} g_i(y,w).\]
This implies that $g_i(y,w)\geq 0$ for every $y\in\calH\cap[\uly,\oly_{i:\uly}]$, in other words, the intersection of the lower $i$-th face of the hyperrectangle $[\uly,\oly]$ and $\calH$.
Similarly, $g_i(y,w) \leq 0$ for every $y\in\calH\cap[\uly_{i:\oly},\oly]$, in other words, the intersection of the upper $i$-th face of the hyperrectangle $[\uly,\oly]$ and $\calH$. 
Thus, for every $y\in\calH\cap \partial[\uly,\oly] = \partial(\calH \cap [\uly,\oly])$, the vector field $g(y,w)$ points into the set $[\uly,\oly]$. But, by Proposition~\ref{prop:invsubspace}, we know that $\calH$ is forward invariant, thus, for every $y\in\partial(\calH\cap[\uly,\oly])$ the vector field $g(y,w)$ points into the set $\calH\cap[\uly,\oly]$. Thus, by Nagumo's theorem~\citet[Theorem 3.1]{FB:99}, the closed set $\calH\cap[\uly,\oly]$ is $[\ulw,\olw]$-robustly forward invariant.
\end{proof}

\newpage
\section{Experiment Implementation Details}

\subsection{Segway Model}

\paragraph{Disturbance Set Partitioning}

To handle the multiplicative disturbance $(1 + w_k)$ for each system parameter from~\eqref{eq:segway}, with $w\in[-0.02,0.02]^{11}$, we partition the disturbance set into $2^{11} = 2048$ different regions, \emph{i.e.}, 
\begin{align*}
    \mathbf{W} = \{\calW^j = W_1\times \cdots\times W_{11} : W_i\in\{[-0.02,0],[0,0.02]\}\}.
\end{align*}
One can create a valid embedding system for the whole disturbance set $\calW = [-0.02,0.02]^{11}$ by considering the worst case for each of the disturbance partitions on each output of the embedding system's vector field,
\begin{align*}
    \begin{aligned} 
        \dot{\uly}_i &= \big(\ul{\sfG}(\ul{\calI}_\calH(\uly,\oly_{i:\uly}),\ol{\calI}_\calH(\uly,\oly_{i:\uly}),\ulw,\olw)\big)_i = \min_{[\ulw^j,\olw^j]\in\bfW} \big(\ul{\sfG}(\ul{\calI}_\calH(\uly,\oly_{i:\uly}),\ol{\calI}_\calH(\uly,\oly_{i:\uly}),\ulw^j,\olw^j)\big)_i, \\
        \dot{\oly}_i &= \big(\ol{\sfG}(\ul{\calI}_\calH(\uly_{i:\oly},\oly),\ol{\calI}_\calH(\uly_{i:\oly},\oly),\ulw,\olw)\big)_i = \max_{[\ulw^j,\olw^j]\in\bfW} \big(\ol{\sfG}(\ul{\calI}_\calH(\uly_{i:\oly},\oly),\ol{\calI}_\calH(\uly_{i:\oly},\oly),\ulw^j,\olw^j)\big)_i.
    \end{aligned}
\end{align*}
These embedding system and $\min$/$\max$ evaluations are vectorized using JAX for efficient evaluation on the GPU.

\paragraph{Training Setup}
\begin{itemize}
    \item Network: We train a $2$ hidden layer network with $32$ neurons each, $3\times 32\times 32\times 1$ with $\relu$ activations.
    \item Data Loss: At each step we uniformly sample $1000$ points $\{x_k\}$ from the set $[-\frac{\pi}{2},\frac{\pi}{2}]\times [-5,5]\times [-2\pi,2\pi]$. We build
    \begin{align*}
        \calL^{\text{data}}(\pi) = \frac1N \sum_{k=1}^N \|\pi(x_k) - Kx_k\|_2^2.
    \end{align*}
    \item Polytope Loss: We use the loss $\calL^\calS$~\eqref{eq:safeloss} with $\lambda = 1000$ and $\varepsilon=0.1$.
    \item Optimizer: Algorithm~\ref{alg:training} terminates in $595$ steps of ADAM with step size $0.001$.
\end{itemize}

\subsection{Platoon of Vehicles with Nonlinearities and Disturbances}

\paragraph{Training Setup}
\begin{itemize}
    \item Network: We train the shared policy $\pi$ as a $3$ hidden layer network with $32$ neurons each, $6\times 32\times 32\times 32\times 1$ with $\relu$ activations.
    \item Data Loss: We use no data loss for this example, \emph{i.e.}, $\calL^{\text{data}} = 0$.
    \item Polytope Loss: We use the loss~\eqref{eq:safeloss} with $\lambda = 1$ and $\varepsilon = 0.02$.
    \item Optimizer: Algorithm~\ref{alg:training} terminates in $\{724,725,807,890,1267\}$ steps of ADAM with step size $0.001$ for platoons with $\{4,10,16,22,28\}$ vehicles.
\end{itemize}

\end{document}